\documentclass{article}

\usepackage{microtype}
\usepackage{graphicx}
\usepackage{subcaption}
\usepackage{booktabs} 

\usepackage{hyperref}\usepackage[table]{xcolor}

\usepackage[preprint]{icml2026}

\usepackage{amsmath}
\usepackage{amssymb}
\usepackage{mathtools}
\usepackage{amsthm}
\usepackage{enumitem}
\usepackage{caption}
\usepackage{ragged2e} 

\usepackage[dvipsnames]{xcolor}
\usepackage[table]{xcolor}
\usepackage[most]{tcolorbox}
\newtcolorbox{mybox}[1][]{%
  enhanced,
  width=\linewidth,
  top=7pt,
  bottom=2pt,
  colback=blue!2,
  colframe=black,
  colbacktitle=black,
  center,
  attach boxed title to top left={yshift=-0.1in,xshift=0.15in},
  boxed title style={boxrule=0pt, colframe=white},
  before upper={\setlength{\parskip}{0.5em}},
  #1
}

\definecolor{HeaderGray}{RGB}{235,235,235}
\definecolor{TokenBlue}{RGB}{210,230,255}
\definecolor{TokenBlueBorder}{RGB}{120,170,240}
\definecolor{search_token}{RGB}{11, 118, 160}
\definecolor{error}{RGB}{234, 76, 72}

\newtcbox{\bluebox}{on line,
  colback=TokenBlue, colframe=TokenBlueBorder,
  boxrule=0.6pt, arc=1.2pt,
  left=1.2pt, right=1.2pt, top=0.8pt, bottom=0.6pt
}

\newtcolorbox{sectionbar}{
  colback=HeaderGray, colframe=black,
  boxrule=0.6pt, arc=0pt,
  left=6pt, right=6pt, top=4pt, bottom=4pt,
  boxsep=0pt
}

\newtcolorbox{examplepanel}{
  colback=white, colframe=black,
  boxrule=0.8pt, arc=0pt,
  left=8pt, right=8pt, top=8pt, bottom=8pt
}

\newtcbox{\ansbox}{on line,
  colback=white, colframe=black,
  boxrule=0.6pt, arc=0pt,
  left=4pt, right=4pt, top=2pt, bottom=2pt
}

\tcbuselibrary{listings,breakable,skins}

\usepackage{fancyvrb}
\usepackage{microtype}
\usepackage{listings}
\usepackage{inconsolata} 
\usepackage{wrapfig}
\usepackage{placeins} 

\lstdefinelanguage{json}{
  basicstyle=\ttfamily\small,
  showstringspaces=false,
  breaklines=true,
  breakatwhitespace=true,
  columns=fullflexible
}

\tcbset{
  mypromptbox/.style={
    enhanced,
    breakable,
    colback=blue!2,          
    colframe=blue!70!black,  
    boxrule=0.8pt,           
    arc=8pt,
    left=8pt,right=8pt,top=6pt,bottom=6pt,
    fonttitle=\bfseries\large,
    coltitle=white,
    colbacktitle=blue!80!black
  }
}

\usepackage[capitalize,noabbrev]{cleveref}
\tcbuselibrary{breakable}
\theoremstyle{plain}

\theoremstyle{definition}

\theoremstyle{remark}

\usepackage[textsize=tiny]{todonotes}
\usepackage{tabularx} 
\usepackage{booktabs}
\usepackage{graphicx}
\usepackage{kotex}
\usepackage{graphicx}
\usepackage{multirow}
\usepackage{colortbl}
\usepackage{makecell}
\usepackage{pifont}
\usepackage[all]{hypcap}
\newcolumntype{C}[1]{>{\centering\arraybackslash}p{#1}}

\icmltitlerunning{Knowledge Integration Decay in Search-Augmented Reasoning of LLMs}

\begin{document}

\twocolumn[
  \icmltitle{Knowledge Integration Decay in\\Search-Augmented Reasoning of Large Language Models}



  \icmlsetsymbol{equal}{*}

  \begin{icmlauthorlist}
    \icmlauthor{Sangwon Yu}{snu}
    \icmlauthor{Ik-hwan Kim}{snu}
    \icmlauthor{Donghun Kang}{snu}
    \icmlauthor{Bongkyu Hwang}{sds}
    \icmlauthor{Junhwa Choi}{sds} \\
    \icmlauthor{Suk-hoon Jung}{sds}
    \icmlauthor{Seungki Hong}{sds}
    \icmlauthor{Taehee Lee}{sds}
    \icmlauthor{Sungroh Yoon}{snu,ipai}
  \end{icmlauthorlist}

  \icmlaffiliation{snu}{Department of Electrical and Computer Engineering, Seoul National University, Seoul, Korea}
  \icmlaffiliation{sds}{Samsung SDS, Korea}
  \icmlaffiliation{ipai}{Interdisciplinary Program in Artificial Intelligence, Seoul National University, Seoul, Korea}

  \icmlcorrespondingauthor{Sungroh Yoon}{sryoon@snu.ac.kr}

  \icmlkeywords{Machine Learning, ICML}

  \vskip 0.3in
]



\printAffiliationsAndNotice{}  

\begin{abstract}
Modern Large Language Models (LLMs) have demonstrated remarkable capabilities in complex tasks by employing search-augmented reasoning to incorporate external knowledge into long chains of thought.
However, we identify a critical yet underexplored bottleneck in this paradigm, termed Knowledge Integration Decay (KID).
Specifically, we observe that as the length of reasoning generated before search grows, models increasingly fail to integrate retrieved evidence into subsequent reasoning steps, limiting performance even when relevant information is available.
To address this, we propose Self-Anchored Knowledge Encoding (SAKE), a training-free inference-time strategy designed to stabilize knowledge utilization.
By anchoring retrieved knowledge at both the beginning and end of the reasoning process, SAKE prevents it from being overshadowed by prior context, thereby preserving its semantic integrity.
Extensive experiments on multi-hop QA and complex reasoning benchmarks demonstrate that SAKE significantly mitigates KID and improves performance, offering a lightweight yet effective solution for knowledge integration in agentic LLMs.
\end{abstract}

\section{Introduction}\label{sec:intro}

The paradigm of Large Language Models (LLMs) has recently undergone a fundamental shift towards test-time scaling, where models leverage extended inference time to solve increasingly complex reasoning tasks~\cite{jaech2024openai, QwQ, guo2025deepseek, yang2025qwen3technicalreport}.
By explicitly generating long chain-of-thought content, modern LLMs have demonstrated remarkable performance gains across domains such as mathematics, programming, and multi-step logical reasoning.
Beyond relying solely on parametric knowledge, this paradigm has been further extended by integrating retrieval-augmented generation, enabling models to perform reasoning grounded in external knowledge sources~\cite{lewis2020rag, jiang-etal-2023-flare, trivedi-etal-2023-ircot}.
By interleaving reasoning with search, contemporary LLMs can dynamically acquire missing information and incorporate it into their inference process, forming the foundation of today's agentic LLM systems~\cite{openaideepresearch, Search-o1, wu2025webdancer, team2025tongyi}.

Despite the promise of this search-augmented reasoning approach, we identify a critical, previously underexplored bottleneck that substantially limits its effectiveness.
Specifically, we observe that a model's ability to incorporate external knowledge is highly sensitive to the length of the reasoning trace generated prior to the search, which we refer to as ``pre-search reasoning''.
As the volume of pre-search reasoning increases, the model's capacity to integrate newly retrieved information degrades sharply.
We term this phenomenon \textbf{Knowledge Integration Decay (KID)}.

Through extensive empirical analysis, we uncover a striking pattern: even when the searched knowledge explicitly contains the correct answer, models frequently fail to utilize it when it is presented after a long reasoning trace.
Moreover, our mechanistic analysis of attention dynamics reveals that the model exhibits a strong bias towards attending to previous reasoning tokens during knowledge encoding.
This issue becomes increasingly pronounced as the reasoning context grows longer.
As a result, the representation of external knowledge becomes distorted by the accumulated state of prior reasoning, preventing faithful integration into the subsequent reasoning process.
The semantic integrity of external knowledge is compromised, as the model treats it merely as a subordinate extension of its existing thoughts rather than as an objective anchor for new insights.

\begin{figure*}[!t]
    \centering
    \includegraphics[width=\textwidth]{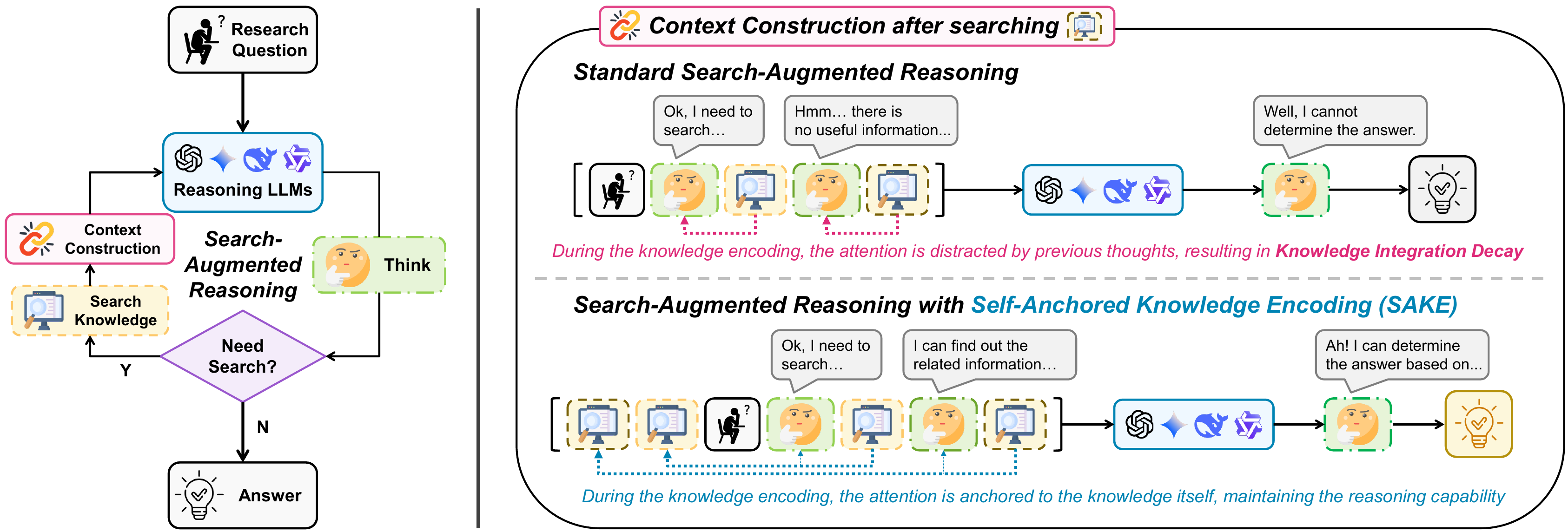} 
    \caption{
    Overview of search-augmented reasoning and post-search context construction.
While standard reasoning suffers from attention interference caused by prior reasoning during knowledge encoding, leading to Knowledge Integration Decay, SAKE anchors attention to searched knowledge, enabling stable and effective reasoning.
    }
    \vspace{-0.3cm}
    \label{fig_method}
\end{figure*}

To address this limitation, we argue that effective search-augmented reasoning requires semantically unbiased encoding of external knowledge.
This intuition is motivated by human reasoning: when humans acquire new information, they interpret it objectively to verify or adjust their hypotheses, rather than forcing it to fit their pre-existing thought.
At the same time, we recognize that simply isolating knowledge from the context is counterproductive.
Essential contextual signals derived from the reasoning process, such as the rationale for the search, the specific intent behind the query, and the strategic relevance of the retrieved information, must be preserved to guide the problem-solving process.
Thus, effective deep reasoning requires a careful balance between preserving the semantic integrity of the knowledge and maintaining its contextual coherence.

To this end, we propose \textbf{Self-Anchored Knowledge Encoding (SAKE)}, a simple yet effective training-free inference-time method.
SAKE is designed to preserve the semantic integrity of external knowledge while retaining necessary reasoning cues.
Instead of merely appending retrieved documents after the reasoning trace, SAKE places the same knowledge both before and after the reasoning process.
This dual placement allows the model to first construct a robust, intrinsic representation of the evidence before it becomes entangled with the specific reasoning path.
When the knowledge reappears later, the attention mechanism naturally anchors back to this earlier representation, effectively neutralizing the interference from pre-search reasoning biases.

We evaluate SAKE extensively on multi-hop QA benchmarks and knowledge-intensive complex reasoning benchmarks, including HotpotQA~\cite{yang2018hotpotqa}, 2WikiMultiHopQA~\cite{ho-etal-2020-2wiki}, MuSiQue~\cite{trivedi2022musique}, FRAMES~\cite{krishna2025frames}, and GAIA~\cite{mialon2024gaia}.
Across all settings, SAKE consistently and substantially improves reasoning performance, achieving gains of up to 37.6\%p over standard search-augmented baselines.
Further analysis confirms that these improvements stem directly from SAKE's ability to mitigate KID by stabilizing knowledge encoding. 

By addressing a fundamental limitation inherent in search-augmented reasoning, SAKE provides a principled and lightweight solution for building more reliable, faithful, and robust agentic LLMs in practice.
Our contributions are summarized as follows:
\begin{itemize}[nosep,leftmargin=*]
    \item We identify \textbf{Knowledge Integration Decay (KID)}, a fundamental limitation in search-augmented reasoning where the ability to incorporate retrieved knowledge degrades as prior reasoning grows longer.
    \item We provide a mechanistic analysis of attention dynamics, showing that excessive attention to previous reasoning distorts the encoding of newly retrieved knowledge.
    \item We propose \textbf{Self-Anchored Knowledge Encoding (SAKE)}, a simple yet powerful inference-time method that preserves semantic integrity while integrating essential reasoning context.
    \item We demonstrate consistent and significant performance gains across multiple multi-hop and knowledge-intensive benchmarks, validating SAKE as an effective solution for mitigating KID.
\end{itemize}

\section{Related Works}\label{sec:related_works}

The fusion of Chain-of-Thought (CoT)~\cite{wei2022chain, kojima2022zeroshot} and test-time scaling~\cite{snell2025scaling, zhang2025surveytesttimescalinglarge} has established inference-time compute as a critical scaling dimension alongside parameter growth~\cite{kaplan2020scalinglawsneurallanguage}.
This shift has given rise to Large Reasoning Models (LRMs)~\cite{xu2025largereasoningmodelssurvey, jaech2024openai, guo2025deepseek, QwQ}, which leverage extended inference capabilities to exhibit System 2 reasoning behaviors, such as planning and self-correction~\cite{muennighoff-etal-2025-s1, li202512surveyreasoning}.
Consequently, these models have achieved state-of-the-art proficiency in complex reasoning domains~\cite{yang2025qwen3technicalreport, he2025skywork}.

To move beyond reliance on fixed end, static parametric knowledge, recent efforts have sought to improve LLM reliability by integrating external knowledge sources via Retrieval-Augmented Generation (RAG)\cite{Huang_2025hallucination, shao2023iterretgen, asai2024selfrag, gao2025smartrag, yu2024autorag}.
Building upon this foundation, search-augmented reasoning has emerged as a more advanced framework.
This approach utilizes test-time scaling to perform deep reasoning interleaved with iterative search~\cite{Search-o1, jin2025searchr1, song2025r1, Li2025WebThinker, hu-etal-2025-webcot}, allowing models to dynamically acquire information as needed.
Furthermore, this paradigm has naturally evolved into modern agentic LLM systems, which employ a diverse set of tools to solve complex, long-horizon, and real-world tasks~\cite{mialon2024gaia, phan2025humanity, wei2025browsecompsimplechallengingbenchmark}. By leveraging search, planning, and tool-use capabilities, recent agentic systems have demonstrated robust performance across a wide range of challenging scenarios~\cite{geminideepresearch, openaideepresearch, zheng-etal-2025-deepresearcher, shi2025deepresearchsystematicsurvey, mei2025aisearchplannermodularagenticsearch, wu2025webdancer, li2025websailornavigatingsuperhumanreasoning}.

Despite the advancements in these works, they largely share a common structural pattern: reasoning is performed first, followed by search, and the retrieved results are subsequently appended to the ongoing reasoning context.
In this work, we identify a fundamental limitation inherent to this sequential reasoning-search-reasoning structure and propose an effective methodology to address it.
Our approach complements and strengthens existing research on search-augmented and agentic LLMs, offering a principled direction for building more reliable and robust reasoning systems.

\section{Knowledge Integration Decay}\label{sec:problem}

\subsection{Preliminaries and Problem Formulation}\label{sec:problem:preliminaries}

We formally define the search-augmented reasoning paradigm commonly adopted by modern reasoning-focused LLMs.
Unlike standard RAG, which typically performs retrieval only once, these models interleave reasoning and search iteratively over multiple steps.

Let $f_\theta$ denote a language model, $q_0$ the input query, and $I$ a task instruction.
At the first step, the model generates an initial reasoning trace:
\begin{equation}
r_1 = f_\theta(I, q_0).
\end{equation}
At each subsequent step $t$, the model issues a search query $q_t$ contained in the current reasoning $r_t$ and retrieves external knowledge $d_t$ using search tool $\mathcal{S}$:
\begin{equation}
d_t = \mathcal{S}(q_t), \quad
q_t \in r_t
\end{equation}
which is then incorporated into the model context for the next reasoning step:
\begin{equation}
r_{t+1} = f_\theta(q_0, r_1, d_1, \dots, r_t, d_t).
\end{equation}
After $K$ steps, the full reasoning context is denoted as:
\begin{equation}
X_K = [I; q_0; r_1; d_1; \dots; r_K; d_K],
\end{equation}
and the final reasoning including the answer prediction is generated as $y = f_\theta(X_K)$.

While this iterative reasoning-retrieval framework enables complex problem-solving, it also introduces a critical challenge: as the reasoning trace grows longer, newly retrieved evidence must be integrated into an increasingly extensive and potentially biasing context.
In the following sections, we demonstrate that this leads to a systematic degradation in knowledge utilization.

\subsection{Empirical Observation}\label{sec:problem:observation}

\begin{figure}[t]
{
\begin{center}
\centerline{\includegraphics[width=\columnwidth]{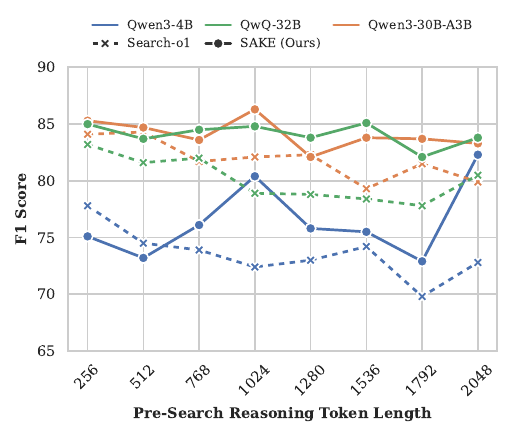}}
\caption{Performance on 2WikiMultihopQA as a function of pre-search reasoning length under oracle search, illustrating Knowledge Integration Decay and the mitigating effect of SAKE.}
\label{fig:problem:observation}
\end{center}
}
\vspace{-1cm}
\end{figure}

We investigate the following research question:

\emph{RQ: Can an LLM consistently maintain its ability to understand and utilize retrieved search results throughout a prolonged reasoning process?}

To answer this question, we empirically study the effect of \emph{pre-search reasoning length} on the performance of search-augmented reasoning models.
Specifically, our goal is to examine whether increasing the volume of reasoning generated before the search degrades the model’s ability to effectively incorporate external knowledge, even when the retrieved information is sufficient to answer the question.

\vspace{-0.3cm}

\paragraph{Experimental Setup}
We conduct experiments on the 2WikiMultihopQA benchmark under a controlled, idealized search setting.
Specifically, we consider a scenario in which the model first performs a sequence of reasoning steps, invokes the search tool once, and then derives the final answer based on the retrieved evidence.
The search-augmented reasoning process is implemented through a prompting-based framework in which the model explicitly decides when to issue a search query during reasoning.
Following prior work~\cite{Search-o1}, we refer to this framework as \textit{Search-o1} throughout this paper.
The prompt template for Search-o1 is provided in Appendix~\ref{sec:appen:prompts}.

We evaluate three representative open-source large reasoning models operating under the Search-o1 framework: \texttt{Qwen3-4B-Thinking}, \texttt{Qwen3-30B-A3B-Thinking}, and \texttt{QwQ-32B}~\cite{QwQ, yang2025qwen3technicalreport}.
To isolate the effect of reasoning length on knowledge integration, we vary the length of the pre-search reasoning trace by controlling the token count of the reasoning generated before the first search step, $r_1$.
After the search is triggered, we ensure that the search result $d_1$ includes all gold supporting documents for the question.
This eliminates retrieval quality as a confounding factor, allowing us to directly measure how effectively the model integrates external knowledge into its reasoning process.
Performance is measured by the word-level F1 score between predictions and ground truth.
Additional details on setup are provided in Appendix~\ref{sec:appen:exp_details}.

\vspace{-0.3cm}

\paragraph{Observation}
Figure~\ref{fig:problem:observation} illustrates the performance trend as the length of pre-search reasoning increases.
Across all Search-o1 baseline models, we observe a consistent degradation in F1 scores as the amount of reasoning performed before search increases.
Notably, this degradation occurs despite all relevant documents being explicitly included in the search results, indicating that the performance drop cannot be attributed to retrieval failure.

Instead, the results suggest that the model increasingly fails to effectively incorporate external knowledge into its subsequent reasoning as the prior reasoning context becomes longer.
This reveals a systematic phenomenon, which we term \textbf{K}nowledge \textbf{I}ntegration \textbf{D}ecay (\textbf{KID}).


We hypothesize that this occurs because extensive earlier reasoning traces $r_1$ tend to dominate the model’s attention during the encoding of $d_1$, thereby biasing its internal representations.
As a result, newly retrieved evidence is overshadowed and consequently underutilized—or even ignored—during final answer generation.
This motivates the mechanistic analysis in the following section, where we examine how attention dynamics contribute to KID.

\subsection{Mechanistic Analysis}\label{sec:problem:analysis}

To understand the underlying cause of Knowledge Integration Decay (KID), we analyze how pre-search reasoning interferes with the encoding of searched knowledge at the attention level.
Specifically, we investigate whether the reasoning context generated before the search disrupts the model's ability to properly encode and represent newly retrieved evidence.

\vspace{-0.3cm}

\paragraph{Quantifying Attention Interference}
Let $A^{l,h} \in \mathbb{R}^{T \times T}$ denote the attention weight matrix of the $h$-th attention head in the $l$-th model layer, where $A^{l,h}_{ij}$ represents the attention weight from token $i$ to token $j$.
For retrieved documents $d_t$ and pre-search reasoning $r_t$, let their token indices range over $[s_{d_t}, e_{d_t}]$ and $[s_{r_t}, e_{r_t}]$, respectively. 

We define the \emph{interference strength} of pre-search reasoning during knowledge encoding as the relative attention mass that tokens in $d_t$ allocate to tokens in $r_t$, compared to the attention allocated within $d_t$ itself:
\begin{equation}
\mathcal{I}^{l,h}(d_t, r_t)
=
\frac{1}{|d_t|}
\sum_{i=s_{d_t}}^{e_{d_t}}
\log
\frac{
\sum_{j=s_{r_t}}^{e_{r_t}} A^{l,h}_{ij}
}{
\sum_{j=s_{d_t}}^{e_{d_t}} A^{l,h}_{ij}
}.
\label{eq:interference_strength}
\end{equation}

This metric quantifies the extent to which the representation of retrieved knowledge is influenced by prior reasoning tokens during its encoding.
A higher value indicates stronger interference, implying that the model allocates significantly more attention to the pre-search reasoning $r_t$ than to the content of the document $d_t$ itself.

We compute this metric across all attention heads and layers and report the average interference strength.
This analysis follows the same experimental setup described in Section~\ref{sec:problem:observation}, utilizing the 2WikiMultihopQA benchmark.
We focus our analysis on the \texttt{Qwen3-4B-Thinking} model to closely examine the fine-grained attention dynamics.

\begin{figure}[t]
{
\begin{center}
\centerline{\includegraphics[width=\columnwidth]{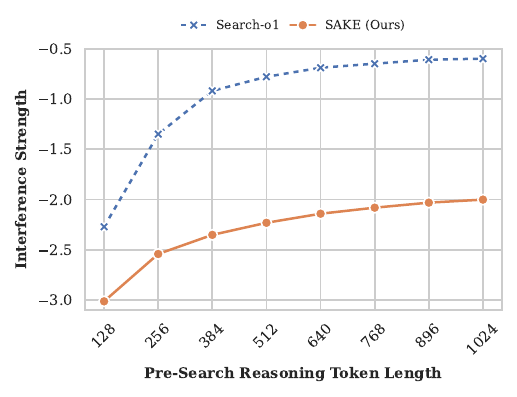}}
\caption{Interference strength as a function of pre-search reasoning length, showing that longer reasoning induces stronger attention-level interference in knowledge encoding.}
\label{fig:problem:attn_noise}
\end{center}
}
\vspace{-0.9cm}
\end{figure}

\vspace{-0.3cm}

\paragraph{Effect of Reasoning Length}
Figure~\ref{fig:problem:attn_noise} presents the interference strength as a function of pre-search reasoning length for the \texttt{Qwen3-4B-Thinking} model.
Under the Search-o1 baseline, we observe a sharp rise in interference strength as the amount of pre-search reasoning increases.
This trend indicates that the retrieved knowledge is increasingly encoded under the heavy influence of prior reasoning, preventing it from being processed as an objective piece of evidence.

This finding provides a mechanistic explanation for the KID phenomenon observed in Section~\ref{sec:problem:observation}.
As the pre-search reasoning grows longer, the attention distribution during knowledge encoding becomes increasingly dominated by earlier reasoning tokens.
This leads to distorted or weakened representations of the newly retrieved information.
Even when relevant evidence is available, the model fails to faithfully integrate it into the subsequent reasoning process.

Taken together, these findings indicate that KID arises from excessive attention-level interference during knowledge encoding.
This insight directly motivates our approach to explicitly regulate this interaction.

Additionally, Figures~\ref{fig:problem:observation} and~\ref{fig:problem:attn_noise} demonstrate that our proposed method, SAKE, substantially mitigates this issue.
A detailed discussion of this effect is provided in Section~\ref{sec:experiments}.

\section{Method}\label{sec:method}


\subsection{Approach of Unbiased Knowledge Encoding}
\label{sec:method:approach}

Building on the empirical observations and mechanistic analysis presented in Section~\ref{sec:problem}, we identify the primary driver of Knowledge Integration Decay (KID) as the excessive interference from the reasoning trace ($r_t$) during the encoding of newly retrieved knowledge ($d_t$).

To address this issue, we introduce \emph{unbiased knowledge encoding} as a core design principle.
This objective is motivated by a simple intuition: when humans consult an external document while solving a problem, they do not distort the factual content based on their partial reasoning states.
Instead, they first comprehend the information as-is and only then determine how it should influence their subsequent reasoning steps.
In contrast, search-augmented LLMs often conflate these two stages, allowing intermediate reasoning states to interfere with and potentially distort the encoding of external evidence.

Formally, we aim to prevent the conditioned representation $g_\theta(d_t \mid r_t)$ from deviating significantly from its intrinsic encoding $g_\theta(d_t)$ due to interference from the reasoning trace.
Importantly, unbiased encoding does not imply that $g_\theta(d_t \mid r_t)$ must be identical to $g_\theta(d_t)$ in isolation.
While the reasoning context should not override or distort the factual content of $d_t$, it still plays a necessary role in shaping how that knowledge is applied.
That is, the reasoning trace should provide functional contextualization without compromising the semantic integrity of the retrieved information.
This allows the encoded knowledge to serve not merely as an isolated fact, but as a context-aware component that effectively supports the next reasoning step $r_{t+1}$.

Accordingly, our objective is not to enforce strict conditional independence between $d_t$ and $r_t$, but rather to suppress harmful interference while preserving beneficial interactions.
We recognize that $r_t$ contains intermediate abstractions essential for problem-solving.
However, its influence should be limited to guiding the usage of knowledge, not altering its meaning.
This principle is summarized as follows:
\begin{enumerate}[nosep,leftmargin=*]
\item \textbf{Semantic Integrity:} External knowledge should be encoded in a manner that preserves its original semantics, ensuring it is not distorted by biases accumulated in prior reasoning steps.
\item \textbf{Reasoning-Aligned Contextualization:} The encoding process should remain aware of the reasoning context, enabling the knowledge to be effectively integrated into the ongoing problem-solving trajectory.
\end{enumerate}
In the following section, we propose a concrete mechanism that operationalizes this principle by explicitly balancing semantic integrity and reasoning contextualization during the knowledge encoding process.

\begin{table*}[!t]
    \centering
    \small
    \caption{Main results (F1 score) on multi-hop QA benchmarks. Compar. denotes Comparison, Infer. denotes Inference, Compos. denotes Compositional, and B. C. denotes Bridge-Comparison questions.}
    \label{tab:mhqa}
        \renewcommand{\arraystretch}{1.2}
        \begin{tabularx}{\textwidth}{l *{9}{>{\centering\arraybackslash}X}}
            \toprule
            \multirow{2}{*}{\textbf{Method}} 
            & \multicolumn{2}{c}{\textbf{HotpotQA}} 
            & \multicolumn{4}{c}{\textbf{2WikiMultiHopQA}} 
            & \multicolumn{3}{c}{\textbf{MuSiQue}} \\
            \cmidrule(lr{2pt}){2-3}
            \cmidrule(lr{2pt}){4-7}
            \cmidrule(lr{2pt}){8-10}
            
            & Compar. & Bridge 
            & Compar. & Infer.& B. C. & Compos.
            & 2-hop & 3-hop & 4-hop \\
            \midrule
            \multicolumn{10}{l}{\textit{\textbf{Qwen3-4B-Thinking}}} \\ 
            Standard RAG                                                & \textbf{48.6}& 47.5& 93.0& 38.6& 56.9& 20.5& 31.5& 21.6&10.9\\ 
            Search-o1                                                & 38.9& 47.6& 91.0& 68.6& 77.0& 48.1& 34.9& 18.8& 9.58
\\ 
            SAKE (Ours)                                                & 47.1& \textbf{53.7}& \textbf{97.0}& \textbf{71.0}& \textbf{86.5}& \textbf{55.0}& \textbf{40.6}& \textbf{24.1}&\textbf{12.0}
\\  \hline
            $\Delta$ (Increased)                                          & + 8.20& + 6.10& + 6.00& + 2.40& + 9.50& + 6.90& + 5.70& + 5.30& + 2.42\\
            \midrule
            \multicolumn{10}{l}{\textit{\textbf{Qwen3-30B-A3B-Thinking}}} \\ 
            Standard RAG                                                & \textbf{39.2}& 48.0& \textbf{92.0}& 58.1& 67.3& 29.1& 32.0& \textbf{22.0}&\textbf{15.8}\\ 
            Search-o1                                                & 20.3& 29.3& 67.9& 46.4& 39.4& 44.0& 25.3& 14.8& 5.94
\\ 
            SAKE (Ours)                                                & 33.3& \textbf{52.4}& 89.2& \textbf{67.5}& \textbf{77.0}& \textbf{50.9}& \textbf{35.2}& 18.0&14.6
\\ \hline
            $\Delta$ (Increased)                                          & + 13.0& + 23.1& + 21.3& + 21.1& + 37.6& + 6.90& + 9.90& + 3.20& + 8.66\\
            \midrule
            \multicolumn{10}{l}{\textit{\textbf{QwQ-32B}}} \\ 
            Standard RAG                                                & \textbf{34.8}& 46.9& \textbf{95.9}& 55.1& 68.9& 24.0& 29.6& 18.4&15.3\\ 
            Search-o1                                                & 26.5& 51.5& 85.3& 69.3& 77.0& 47.0& 35.8& 15.0& 12.4
\\ 
            SAKE (Ours)                                                & 34.2& \textbf{53.6}& 94.8& \textbf{72.6}& \textbf{82.0}& \textbf{52.1}& \textbf{37.6}& \textbf{23.0}&\textbf{20.4}
\\ \hline
            $\Delta$ (Increased)                                          & + 7.70& + 2.10& + 9.50& + 3.30& + 5.00& + 5.10& + 1.80& + 8.00& + 8.00\\ 
            \bottomrule
        \end{tabularx}
\end{table*}

\subsection{SAKE: Self-Anchored Knowledge Encoding}
\label{sec:method:sake}

To mitigate KID while preserving both semantic integrity and reasoning contextualization, we propose \textbf{S}elf-\textbf{A}nchored \textbf{K}nowledge \textbf{E}ncoding (\textbf{SAKE}). 
SAKE introduces a dual-positioning strategy for external knowledge, in which search results are integrated into the input sequence in two complementary roles: (1) a cumulative knowledge stack that serves as a semantic anchor, and (2) a reasoning-aligned context that facilitates stepwise knowledge utilization.

\subsubsection{Cumulative Knowledge Stack as a Semantic Anchor}

The first component of SAKE is a cumulative knowledge stack placed at the very beginning of the reasoning trace.
At reasoning step $t+1$, the model is provided with a stack of all previously searched documents, organized in reverse chronological order:
\begin{equation}
S_t = [d_t, d_{t-1}, \dots, d_1].
\end{equation}
Here, the most recently searched information is placed at the front of the stack, while earlier documents appear later in the sequence.
By positioning $S_t$ before any generated reasoning tokens ($r_1, \dots, r_t$), the model encodes external knowledge in a context, free from interference by prior reasoning.
Furthermore, placing the most recent search result ($d_t$) at the very beginning ensures that it is processed with minimal exposure to preceding content under the causal attention mechanism.

As a result, the document containing the latest searched information is encoded with controlled input sequence construction, preventing speculative reasoning traces from distorting its semantics.
The cumulative stack therefore serves as a \emph{semantic anchor}, preserving the semantic integrity of external knowledge and mitigating the distortion that commonly arises when search results are simply appended after the reasoning.

\subsubsection{Reasoning-Aligned Contextualization}

While the knowledge stack preserves semantic integrity, effective reasoning still requires that knowledge be interpreted in relation to the current problem state.
To this end, SAKE additionally retains the conventional placement of the search result $d_t$ immediately after the corresponding reasoning step $r_t$, which we refer to as reasoning-aligned contextualization.

In this post-reasoning position, $d_t$ is explicitly conditioned on the current reasoning trace, enabling the model to contextualize the knowledge for the ongoing reasoning step $r_{t+1}$.
Because the same documents have already been encoded in the initial knowledge stack ($S_t$), their semantic representation at this later stage is implicitly anchored to the earlier, pristine representation.
This dual exposure allows the model to leverage contextual relevance without allowing the reasoning trace to overwrite or distort the underlying factual content.
A detailed information-theoretic formulation of SAKE is provided in Appendix~\ref{sec:appen:mi_sake}.

\subsubsection{Formal Sequence Formulation}

Formally, the input sequence at the reasoning step $t+1$ is constructed as follows:
\begin{equation}
X_{t+1} =
[I;\, 
\underbrace{[d_t; \dots; d_1]}_{S_t}; q_0;\,
\overbrace{r_1; d_1; \dots; r_t; d_t}^{\text{Reasoning-aligned traces}} ].
\label{eq:formal_seq_formulation}
\end{equation}
This formulation explicitly separates the roles of knowledge preservation and knowledge utilization.
The knowledge stack enforces semantic stability, while the interleaved reasoning-knowledge pairs enable adaptive reasoning.
Through this design, SAKE effectively mitigates KID by balancing robust semantic encoding with contextual flexibility throughout the deep reasoning process.

\section{Experiments}\label{sec:experiments}

\subsection{Experimental Setup}

We evaluate SAKE on a diverse suite of knowledge-intensive benchmarks that demand complex, multi-step reasoning grounded in external knowledge.
Specifically, we adopt three widely established multi-hop QA benchmarks: \textbf{HotpotQA}, \textbf{2WikiMultiHopQA}, and \textbf{MuSiQue}. We also consider the \textbf{FRAMES} benchmark, which poses more challenging multi-hop reasoning scenarios.
In addition, to assess the practical efficacy of our method in realistic, open-ended reasoning scenarios, we perform evaluations on the \textbf{GAIA} benchmark. We report word-level F1 scores on the multi-hop QA benchmarks and LLM-as-a-Judge accuracy on FRAMES and GAIA.

We apply SAKE to three representative reasoning models: \texttt{Qwen3-4B-Thinking}, \texttt{Qwen3-30B-A3B-Thinking}, and \texttt{QwQ-32B}.
For each model, we compare performance against two primary baselines:
\textbf{(1) Standard RAG}, where retrieved documents are simply appended to the input context before the reasoning process begins, and  
\textbf{(2) Search-o1}~\cite{Search-o1}, a representative search-augmented reasoning framework where retrieval steps are dynamically interleaved with intermediate reasoning steps.

For all benchmarks other than GAIA, we build benchmark-specific search corpora and conduct embedding-based retrieval, whereas GAIA is evaluated using web search.

Implementation details, including dataset processing, evaluation protocol, search environments, and generation configuration, are provided in Appendix~\ref{sec:appen:exp_details}.
The prompts used for each method are provided in Appendix~\ref{sec:appen:prompts}.

\subsection{Main Results}

\begin{figure}[t]
    \centering
    \vspace{-8pt}
    \includegraphics[width=0.95\columnwidth]{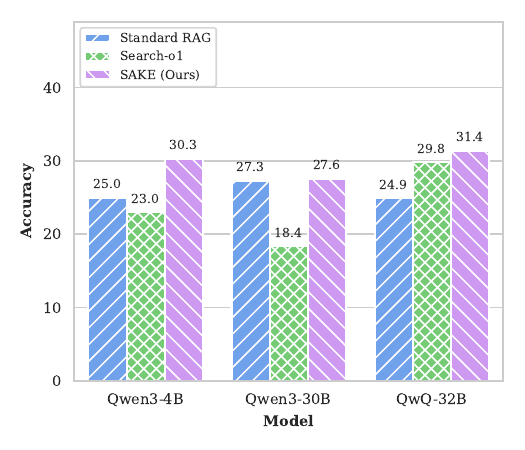}
    \vspace{-6pt}
    \caption{Performance on FRAMES across different models and methods.}
    \label{fig:problem:laj_frames}
    \vspace{-0.3cm}
\end{figure}

Table~\ref{tab:mhqa} presents the comparative performance on the multi-hop QA benchmarks (HotpotQA, 2WikiMultiHopQA, and MuSiQue).
Across all evaluated settings, SAKE consistently outperforms the Search-o1 baseline, demonstrating substantial gains in reasoning accuracy.
Notably, the performance improvement becomes significantly more pronounced as the complexity of the query increases.
For instance, we observe particularly large gains (e.g., +37.6 F1) on bridge-comparison questions.
It is worth noting that the Search-o1 baseline occasionally underperforms even Standard RAG.
This result attributes to Knowledge Integration Decay (KID), where the extended reasoning traces in Search-o1 inadvertently hinder the effective utilization of retrieved evidence.
SAKE reliably reverses this trend and surpasses both baselines, confirming its effectiveness in preserving and utilizing external knowledge even within long reasoning contexts.

Figure~\ref{fig:problem:laj_frames} presents the evaluation results on the FRAMES benchmark across varying model types, computed using an LLM-as-a-judge framework.
As shown, SAKE consistently outperforms the Search-o1 baseline across all three models.
Given that FRAMES requires complex reasoning, including numerical reasoning and multiple constraints, these results suggest that SAKE effectively balances the use of external information with the internal reasoning context, ensuring robust performance even in demanding scenarios.

Table~\ref{tab:gaia} reports the results on the GAIA benchmark, where similar positive trends are observed.
SAKE achieves consistent improvements over both direct generation and Search-o1 baselines.
The performance gap between SAKE and the baselines widens as the task difficulty increases from Level 1 to Level 2.
This suggests that SAKE is particularly effective at maximizing the model’s reasoning potential near its capability boundary, providing robust support for complex, real-world tasks.

\begin{table}[!t]
    \centering
    \small
    \caption{Main results (LLM-as-a-Judge Acc.) on GAIA benchmark.}
    \label{tab:gaia}
        \renewcommand{\arraystretch}{1.15}
        \begin{tabular}{lcccc}
            \toprule
            \multirow{2}{*}{\textbf{Method}} 
            & \multicolumn{4}{c}{\textbf{GAIA-Text}} \\
            \cmidrule(lr{2pt}){2-5}
            & Level 1 & Level 2 & Level 3 & Avg. \\
            \midrule
            \multicolumn{5}{l}{\textit{\textbf{Qwen3-4B-Thinking}}} \\ 
            Direct Gen.                                                 & 30.8& 17.3& 0.00& 20.4
\\ 
            Search-o1                                                   & 48.7& 48.1& 8.33& 43.7
\\ 
            SAKE (Ours)                                                  & \textbf{56.4}& \textbf{55.8}& 8.33& \textbf{50.5}\\
            \midrule
            \multicolumn{5}{l}{\textit{\textbf{Qwen3-30B-A3B-Thinking}}} \\ 
            Direct Gen.                                                 & 30.8& 13.5& 8.33& 19.4
\\ 
            Search-o1                                                   & 56.4& 32.7& 16.7& 39.8
\\ 
            SAKE (Ours)                                                 & \textbf{59.0}& \textbf{42.3}& 16.7& \textbf{45.6}\\
            \midrule
            \multicolumn{5}{l}{\textit{\textbf{QwQ-32B}}} \\ 
            Direct Gen.                                                  & 35.9& 17.3& 8.33& 23.3
\\ 
            Search-o1                                                   & 48.8& 44.2& \textbf{16.7}& 42.7\\ 
            SAKE (Ours)                                                 & \textbf{56.4}& \textbf{57.7}& 8.33& \textbf{51.5}\\
            \bottomrule
        \end{tabular}
\vspace{-0.1cm}
\end{table}

\subsection{Further Analysis}

\paragraph{Mitigation of KID}
Figure~\ref{fig:problem:observation} illustrates the performance trajectory as the length of pre-search reasoning increases.
While the Search-o1 baseline exhibits a clear decline in performance (confirming the presence of KID), SAKE maintains remarkable stability.
This confirms that SAKE effectively insulates the knowledge integration process from the detrimental effects of extended reasoning contexts.

\vspace{-0.3cm}

\paragraph{Reduction of Interference}
Consistent with the performance stability, Figure~\ref{fig:problem:attn_noise} demonstrates that the interference strength, the metric defined in Section~\ref{sec:problem:analysis} to measure attention bias, grows much more slowly under SAKE compared to the baseline.
This indicates that our dual-positioning strategy successfully suppresses excessive attention to prior reasoning during the knowledge encoding phase.

\vspace{-0.3cm}

\begin{figure}[t]
    \centering
    \includegraphics[width=0.82\columnwidth]{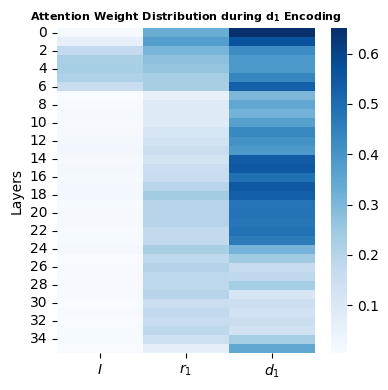}
    \caption{Attention distribution while the model encodes the searched knowledge ($d_1$) through Search-o1.}
    \label{fig:results:attn_map_searcho1}
    \vspace{-0.4cm}
\end{figure}

\paragraph{Mechanism of Action}
To examine attention-level interference in greater detail, we visualize the attention distributions during the encoding of external knowledge, as shown in Figures~\ref{fig:results:attn_map_searcho1} and \ref{fig:results:attn_map_sake}.
A clear contrast emerges between Search-o1 and SAKE: Search-o1 exhibits a high concentration of attention on the prior reasoning context $r_1$, particularly in deeper layers, while allocating substantially weaker attention to the retrieved document $d_1$.
In contrast, SAKE demonstrates a structured shift in attention across layers: early layers partially attend to $r_1$ to capture relevant contextual cues, whereas deeper layers progressively shift the attention toward the knowledge stack $S_1$.
This layer-wise transition enables the model to decouple the semantic encoding of retrieved knowledge from prior reasoning, thereby preserving the semantic integrity of $d_1$, maintaining robust focus on the document content, and effectively mitigating KID.

\section{Discussion}\label{sec:discussion}

\subsection{Is Performance Gain of SAKE Truly due to Mitigating KID?}

A critical question regarding SAKE is whether its performance gains stem from the structural mitigation of KID or simply from the redundancy of having the retrieved content appear twice in the context, as some prior works have suggested regarding repetition~\cite{springer2024repetition, yu2025unleashing}.
To disentangle these factors, we construct a control variant where the retrieved document is simply duplicated immediately following the reasoning step ($[\dots; r_t; d_t; d_t]$) instead of adopting the dual-placement structure of SAKE.

This setting, denoted as ``Repeat'' in Table~\ref{tab:abl_other_settings}, isolates the effect of repetition from SAKE.
This variant yields little to no improvement over the Search-o1 baseline and, in some cases, even degrades performance.
In contrast, SAKE consistently achieves substantial gains.
These results indicate that the effectiveness of SAKE cannot be attributed to simple repetition.
Instead, the gains arise from its ability to mitigate KID by anchoring the knowledge in a position free from interference, thereby ensuring robust encoding.

\subsection{Ablation on Reasoning-Aligned Contextualization}

We further analyze the necessity of the reasoning-aligned contextualization component (the second placement of $d_t$ after $r_t$).
To do this, we evaluate a ``Stack Only'' variant, where retrieved documents are provided only at the beginning of the context (i.e., $[I; d_t; \dots; q_0; \dots r_t]$) and are not reintroduced after the reasoning step.

As shown in Table~\ref{tab:abl_other_settings}, this variant outperforms the vanilla Search-o1 baseline, suggesting that early knowledge injection is indeed beneficial for securing a clean semantic representation.
However, it consistently underperforms the full SAKE method.
This result highlights that semantic integrity alone is insufficient.
Effective search-augmented reasoning requires not only the preservation of factual content but also its proper contextualization within the ongoing reasoning trace to guide the specific problem-solving path.

\begin{figure}[t]
{
\begin{center}
\centerline{\includegraphics[width=\columnwidth]{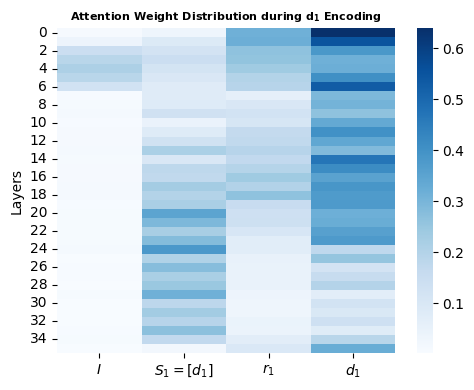}}
\caption{Attention distribution across the input context while the model encodes the searched knowledge ($d_1$) through SAKE.}
\label{fig:results:attn_map_sake}
\end{center}
}
\vspace{-0.9cm}
\end{figure}

\subsection{Why Resolving KID Is Essential for Scaling Search-Augmented Reasoning}

Finally, we investigate a fundamental implication for the field: does enhancing search quality automatically lead to better reasoning?
To answer this, we compare a realistic retrieval setting (“Hard Search”) with an idealized setting where the search space is restricted to a small set of highly relevant documents (“Easy Search”).

\begin{figure}[t]
{
\begin{center}
\centerline{\includegraphics[width=0.9\columnwidth]{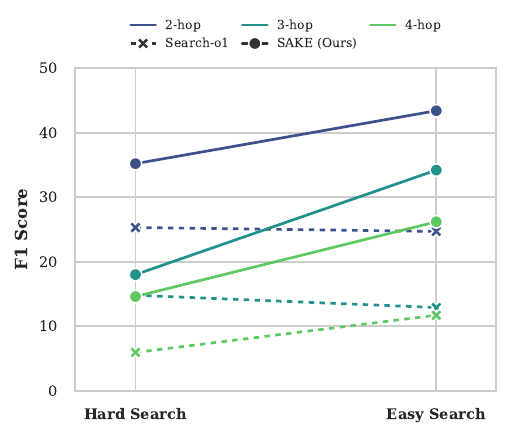}}
\caption{Performance under different search conditions.}
\label{fig:discussion:easy_to_hard_search}
\end{center}
}
\vspace{-0.7cm}
\end{figure}

As shown in Figure~\ref{fig:discussion:easy_to_hard_search}, improving the search environment leads to only marginal gains for the Search-o1 baseline.
This suggests that current models face a ``utilization bottleneck'', meaning that even when perfect information is retrieved, they struggle to use it effectively due to KID.
In contrast, SAKE exhibits substantial performance improvements under the same high-quality search conditions.
Our findings suggest that resolving KID is a prerequisite for fully unlocking the potential of future agentic reasoning systems. Without addressing this integration decay, investments in better retrieval tools or larger context windows may yield diminishing returns.

\section{Conclusion}\label{sec:conclusion}

In this work, we identify \textbf{K}nowledge \textbf{I}ntegration \textbf{D}ecay (\textbf{KID}) as a fundamental yet previously overlooked bottleneck in search-augmented reasoning.
We show that as pre-search reasoning extends, a model's ability to effectively incorporate newly retrieved evidence degrades.
Through mechanistic analyses, we demonstrate that this failure arises from attention-level interference during knowledge encoding, where the accumulated state of prior reasoning overshadows the objective representation of external knowledge.

\begin{table}[!t]
    \centering
    \small
    \caption{Results for SAKE and the variants on MuSiQue.}
    \label{tab:abl_other_settings}
    \resizebox{0.8\columnwidth}{!}{
        \renewcommand{\arraystretch}{1.25}
        \begin{tabular}{lccc}
            \toprule
            \multirow{2}{*}{\textbf{Method}} 
            & \multicolumn{3}{c}{\textbf{MuSiQue}} \\
            \cmidrule(lr{2pt}){2-4}
            
            & 2-hop & 3-hop & 4-hop \\
            \midrule
            Search-o1                                                   &25.3& 14.8& 5.94
\\ 
            SAKE (Ours)                                                 & \textbf{35.2}& \textbf{18.0}& \textbf{14.6}
\\
            Repeat                                                      & 26.6& 10.2&7.10
\\ 
            Stack Only                                                   & 31.4& 16.5&13.2
\\ 
            \bottomrule
        \end{tabular}
    }
\vspace{-0.3cm}
\end{table}


To address this issue, we propose \textbf{S}elf-\textbf{A}nchored \textbf{K}nowledge \textbf{E}ncoding (\textbf{SAKE}), a simple yet effective inference-time strategy.
SAKE is designed to balance two essential requirements: preserving the semantic integrity of retrieved knowledge and maintaining the necessary reasoning contextualization.
Extensive experiments across multi-hop QA and complex reasoning benchmarks confirm that SAKE consistently improves both performance and robustness---particularly as reasoning depth increases---without requiring any additional training or architectural modifications.

\vspace{-0.3cm}

\paragraph{Limitations}
While this work provides a rigorous analysis of KID and a robust mitigation strategy, we acknowledge a practical trade-off in context management, as our dual-placement strategy inherently increases the input length with each search step.
Furthermore, since SAKE operates solely at inference time, it leaves open the possibility for fundamental training-time solutions.
Future work may build upon these findings to explore adaptive context management techniques, integrate KID-aware objectives into reinforcement learning for search policies, and extend these principles to broader applications in autonomous agent systems.

\section*{Impact Statements}
This work advances the understanding of knowledge integration dynamics in Large Language Models (LLMs), identifying Knowledge Integration Decay (KID) as a systematic bottleneck that limits the effectiveness of search-augmented reasoning.
By demonstrating that models often fail to utilize retrieved evidence despite its availability, we highlight a critical reliability issue in current agentic systems.
By proposing a lightweight, training-free mitigation strategy, our research offers a practical path toward building more robust and faithful reasoning systems.
This is particularly impactful for knowledge-intensive applications such as open-domain question answering, scientific research assistance, and autonomous decision-making, where the faithful grounding of arguments in external facts is paramount.

While our approach significantly enhances reasoning reliability, we acknowledge the associated computational trade-offs, specifically the increased inference cost due to the expanded context length required for dual knowledge placement.
Furthermore, as an inference-time intervention, our method addresses the manifestation of KID rather than its root causes in the model training process.
As such, this work does not directly raise new societal risks; rather, it contributes to AI safety by reducing the likelihood of models ignoring factual evidence and hallucinating.
We hope this study serves as a catalyst for future research into training-centric solutions, efficient context management, and the development of trustworthy agents that faithfully ground their reasoning in external reality.

\section*{Acknowledgements}
This work was supported by the Institute of Information \& Communications Technology Planning \& Evaluation (IITP) grants funded by the Korea government (MSIT) [NO.RS-2021-II211343, Artificial Intelligence Graduate School Program (Seoul National University); No.2022-0-00959, RS-2022-II220959], by the National Research Foundation of Korea (NRF) grant [No.2022R1A3B1077720, 2022R1A5A7083908], BK21 FOUR Program of the Education and Research Program for Future ICT Pioneers, Seoul National University in 2026, and by the Samsung SDS Co., Ltd.


\bibliography{icml2026}
\bibliographystyle{icml2026}

\newpage
\appendix
\onecolumn

\section{Information-Theoretic View of Attention Anchoring in SAKE}\label{sec:appen:mi_sake}

We provide a compact information-theoretic interpretation of why Self-Anchored Knowledge Encoding (SAKE) mitigates Knowledge Integration Decay (KID). Let $D$ denote the retrieved document at step $t$ (i.e., $d_t$), $R$ the pre-search reasoning trace (i.e., $r_t$), and $Z$ the remaining context tokens such as instruction/query and other non-target blocks (e.g., $I,q,d_{<t}$). Let $H$ be the hidden representation used to encode tokens in $D$ under the Transformer model with causal self-attention:
\begin{equation}
H \;=\; f_\theta(D \mid R, Z).
\end{equation}
A natural notion of \emph{knowledge preservation} is the mutual information between the encoded representation and the underlying evidence:
\begin{equation}
\mathcal{K} \;\triangleq\; I(H; D).
\end{equation}
Intuitively, KID corresponds to the regime where $I(H;D)$ decreases as the reasoning context grows, meaning the model's representation carries less information about the retrieved evidence even when $D$ is sufficient for answering.

\paragraph{Reasoning interference as conditional dependence.}
Using the chain rule,
\begin{equation}
I(H;D) \;=\; I(H;D \mid R,Z) \;-\; I(H;R,Z \mid D).
\label{eq:mi_decomp}
\end{equation}
The second term, $I(H;R,Z \mid D)$, captures how strongly the encoding of $D$ is \emph{contaminated} by non-evidence context. In standard search-augmented reasoning, where $D$ is appended after a long reasoning trace, causal attention allows tokens in $D$ to attend broadly to $R$; empirically this increases attention-level interference (Eq.~\eqref{eq:interference_strength}, Figure~\ref{fig:problem:attn_noise}), which corresponds to an increase in $I(H;R,Z \mid D)$ and thus a decrease in $I(H;D)$ via Eq.~\eqref{eq:mi_decomp}.

\paragraph{Attention budget and the role of SAKE.}
For a token $x_i \in D$ at some layer/head, denote the total attention mass allocated to a token subset $S$ as
\begin{equation}
\alpha_i(S) \;\triangleq\; \sum_{j \in S} A_{ij},
\end{equation}
where $A_{ij}$ is the causal attention weight. Importantly, $\alpha_i(D)+\alpha_i(R)\neq 1$ in general because the prefix contains additional tokens $Z$. Instead, the relevant quantity is the \emph{relative} allocation within the accessible prefix:
\begin{equation}
\pi_i(R;D,Z) \;\triangleq\; \frac{\alpha_i(R)}{\alpha_i(D)+\alpha_i(R)+\alpha_i(Z)} \;\;\;\; \in [0,1],
\label{eq:rel_mass}
\end{equation}
which increases when attention shifts from evidence tokens to reasoning/context tokens. Since $H$ is a convex mixture of value vectors weighted by $A_{ij}$, larger $\pi_i(R;D,Z)$ implies the representation is more dependent on $(R,Z)$ and less on $D$, increasing $I(H;R,Z \mid D)$ and reducing $\mathcal{K}=I(H;D)$.

SAKE prevents this failure mode by placing $D$ as a \emph{knowledge stack} before reasoning (Eq.~\eqref{eq:formal_seq_formulation}). For the first occurrence of $D$ in the stack, causal masking implies $\alpha_i(R)=0$ for all $x_i \in D$, hence $\pi_i(R;D,Z)=0$ and the encoding satisfies
\begin{equation}
p(H \mid D,R,Z) \;=\; p(H \mid D,Z),
\end{equation}
yielding a strictly smaller interference term $I(H;R \mid D,Z)$ and thus a larger $I(H;D)$ than the standard ordering. When $D$ reappears after reasoning (reasoning-aligned contextualization), the model can leverage $R$ for \emph{utilization} while the earlier stack representation acts as an information-preserving anchor, consistent with the layer-wise attention shift toward the stack observed in Figure ~\ref{fig:results:attn_map_sake}. Overall, SAKE can be viewed as maximizing the mutual information between representations and retrieved evidence by structurally minimizing conditional dependence on the reasoning trace during knowledge encoding.

\paragraph{Why later occurrences of $D$ are anchored to the knowledge stack.}
A remaining question is why the second occurrence of the retrieved document $D$, appearing after the reasoning trace, does not reintroduce severe interference from $R$. Importantly, SAKE does \emph{not} aim to eliminate attention to $R$ because some dependence on $R$ is necessary for reasoning-aligned contextualization. Instead, SAKE reduces the relative influence of $R$ on the representation of $D$ by providing an earlier, reasoning-independent semantic anchor $D_{\text{stack}}$.

Consider a token $x_i \in D$ in the post-reasoning position. Its attention-based update decomposes as
\begin{equation}
h_i = \sum_{j < i} A_{ij} v_j
= \sum_{j \in D_{\text{stack}}} A_{ij} v_j
+ \sum_{j \in R} A_{ij} v_j
+ \sum_{j \in Z} A_{ij} v_j,
\end{equation}
where $D_{\text{stack}}$ denotes the earlier occurrence of the same retrieved knowledge, and $Z$ collects other context tokens (e.g., instruction/query and other blocks).

Because $D_{\text{stack}}$ and the current $D$ share highly similar semantic content, dot-product attention induces stronger query--key alignment between tokens in $D$ and those in $D_{\text{stack}}$ than between $D$ and unrelated reasoning tokens:
\begin{equation}
\mathbb{E}\!\left[q_i^\top k_{D_{\text{stack}}}\right]
\;\;>\;\;
\mathbb{E}\!\left[q_i^\top k_{R}\right],
\end{equation}
which increases the attention mass on $D_{\text{stack}}$ relative to $R$:
\begin{equation}
\sum_{j \in D_{\text{stack}}} A_{ij}
\;\;\gg\;\;
\sum_{j \in R} A_{ij},
\label{eq:anchor_mass}
\end{equation}
This does not imply $\sum_{j \in R} A_{ij}=0$; rather, it ensures that $R$ contributes contextual cues for the reasoning behavior while $D_{\text{stack}}$ provides the dominant semantic basis for encoding $D$.

In information-theoretic terms, introducing $D_{\text{stack}}$ reduces the conditional dependence of the knowledge representation on the reasoning trace:
\begin{equation}
I(H;R \mid D_{\text{stack}}, Z)
\;\;\ll\;\;
I(H;R \mid Z),
\label{eq:mi_reduction}
\end{equation}
so the mutual information between the representation and the evidence is better preserved:
\begin{equation}
I(H;D \mid Z)
\;\;\text{is larger under SAKE than under the standard ordering.}
\end{equation}
This perspective aligns with our empirical observation that attention gradually shifts from $R$ toward the knowledge stack in deeper layers (Figure~\ref{fig:results:attn_map_sake}), indicating that SAKE enables contextualization without semantic overwrite.

\section{Experimental Details}\label{sec:appen:exp_details}

\subsection{Dataset Processing}

\paragraph{Data filtering.}
To ensure that performance improvements reflect enhanced knowledge integration rather than shortcut reasoning from parametric knowledge, we apply a filtering procedure to all evaluation datasets except GAIA.

Specifically, for each benchmark, we first identify \emph{hard samples} that cannot be correctly solved by the model using direct reasoning alone, without access to external knowledge.
For each query in the validation set, we sample 8 independent responses from the model under a no-retrieval setting.
If any of these responses achieves a word-level F1 score greater than 0.5 against the ground-truth answer (for HotpotQA, 2WikiMultiHopQA, and MuSiQue), or is judged as \texttt{Correct} by an LLM-based evaluator (for FRAMES and GAIA), the corresponding sample is considered solvable via parametric knowledge and is excluded from evaluation.

This filtering process ensures that the resulting evaluation set consists solely of samples that genuinely require external knowledge search.
For HotpotQA, 2WikiMultiHopQA, and MuSiQue, we additionally balance the evaluation set to maintain an equal number of samples across query types.
For FRAMES, we randomly sample up to 400 instances from the remaining hard examples.
Table~\ref{tab:data_statistics} summarizes the number of evaluation samples per benchmark and per model after filtering.

For GAIA, we do not apply additional filtering, as its validation set is already small (165 instances) and designed to require complex multi-step reasoning.
However, since our models operate purely in the textual modality, we restrict evaluation to the text-only subset of GAIA, consisting of 103 questions.

\paragraph{Pre-search reasoning length control during KID analysis.}
In the analysis presented in Section~\ref{sec:problem}, we investigate how the length of pre-search reasoning affects KID.
To this end, we explicitly control the amount of reasoning performed before retrieval.

Concretely, we first generate a direct reasoning trace without retrieval, denoted as
\[
r_0 = f_\theta(q),
\]
and truncate it to a fixed length to obtain $\tilde{r}_0$.
The model then continues reasoning from $\tilde{r}_0$ to produce a search-triggering trace,
\[
r_1 = f_\theta(I, q, \tilde{r}_0).
\]

The search query extracted from $r_1$ is used to retrieve documents via an embedding-based retrieval module.
As mentioned in Section~\ref{sec:problem}, we inject gold documents into the retrieved results to construct an idealized search setting, ensuring that retrieval quality does not confound the analysis.
The model is then tasked with producing the final answer based on these retrieved documents.

\begin{table}[!t]
    \centering
    \small
    \caption{Data statistics for each evaluation set.}
    \label{tab:data_statistics}
        \renewcommand{\arraystretch}{1.25}
        \begin{tabular}{lccccc}
            \toprule
            {\textbf{Model}} & \textbf{HotpotQA} & \textbf{2WikiMultiHopQA} &\textbf{MuSiQue} & \textbf{FRAMES} & \textbf{GAIA} \\
            \midrule
            Qwen3-4B         & 400& 400& 450& 400&                                      
103\\ 
            Qwen3-30B-A3B     & 400& 400& 450& 370&                                        
103\\
            QwQ-32B   & 400& 400&450& 389&
103\\ 
            \bottomrule
        \end{tabular}
\end{table}


\subsection{Evaluation Metrics}

We evaluate model performance using two complementary metrics: word-level F1 score and LLM-as-a-Judge accuracy, depending on the characteristics of each benchmark. For HotpotQA, 2WikiMultiHopQA, and MuSiQue, the ground-truth answers are provided as short, span-based responses derived from the gold documents.
Accordingly, we adopt the word-level F1 score between the model-generated answer and the ground-truth answer as the evaluation metric, following standard practice in multi-hop question answering.

In contrast, FRAMES and GAIA often involve long-form or compositional answers, where semantically correct responses may not exactly match the reference text.
Moreover, these benchmarks frequently require reasoning over implicit or multi-step evidence, making exact string matching unreliable.
To account for this, we employ an LLM-based evaluation protocol, using \texttt{Qwen3-30B-A3B-Instruct} as the judge model to assess answer correctness.
The judge determines whether the model output is semantically consistent with the ground-truth answer, following the evaluation setup described in prior work.

\subsection{Implementation}

\paragraph{Search environment.}
For HotpotQA, 2WikiMultiHopQA, and MuSiQue, we construct the search corpus by aggregating all document chunks provided in the validation split of each benchmark.
This results in 66,635 documents for HotpotQA, 56,687 for 2WikiMultiHopQA, and 101,962 for MuSiQue.

For FRAMES, each sample is associated with a relevant Wikipedia page rather than pre-segmented document chunks.
Accordingly, we extract the main textual content of each page up to a maximum length of 8,000 characters and segment it into paragraph-level chunks.
This procedure yields a total of 25,106 paragraphs.
Because these paragraphs are generally longer than the document chunks used in other benchmarks, the effective amount of information retrieved per search step is larger for FRAMES.

All corpora are encoded using \texttt{Qwen3-Embedding-0.6B} to construct an embedding-based retrieval index.
During search-augmented reasoning, the model retrieves the top-5 most relevant documents at each search step and is allowed to perform up to 10 search iterations.
For the Standard RAG baseline, the model retrieves the top-10 documents only once at the beginning of inference.

For GAIA, which requires open-domain information gathering, we employ a web search tool powered by \texttt{GPT-4.1-mini}.
During search-augmented reasoning, the generated search queries are issued to \texttt{GPT-4.1-mini}, which performs web search and returns the retrieved results to be incorporated into subsequent reasoning steps.

\paragraph{Generation details.}
We use identical decoding hyperparameters for all evaluated models.
Specifically, responses are generated with temperature $0.6$, top\_$p$ $0.95$, top\_$k$ $20$, and min\_$p$ $0.0$.

For search-augmented reasoning, the model is allowed to generate up to 4,096 tokens per reasoning step, with a maximum of 30,000 tokens per example.
For GAIA, which involves longer and more complex reasoning chains, we increase the limit to 8,192 tokens per step and 50,000 tokens in total.
For direct reasoning and standard RAG baselines, the maximum generation length is set to 16,384 tokens.

\section{Additional Results}\label{sec:appen:add_exp}

\subsection{Additional Analysis on KID}
\begin{figure*}[t]
    \centering
    \begin{subfigure}[b]{0.48\textwidth}
        \centering
        \includegraphics[width=\linewidth]{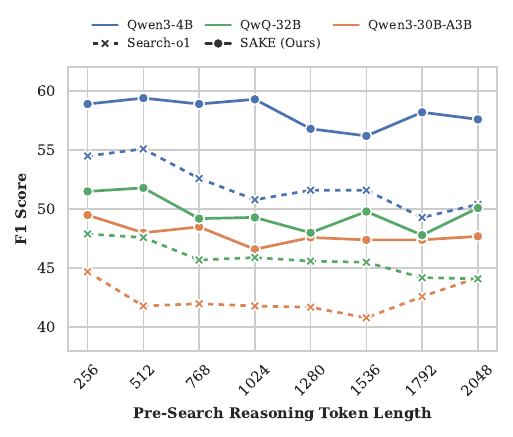}
        \caption{HotpotQA performance across varying pre-search reasoning lengths using oracle search.}
        \label{fig:problem:kid_result_hotpot}
    \end{subfigure}
    \hfill 
    \begin{subfigure}[b]{0.48\textwidth}
        \centering
        \includegraphics[width=\linewidth]{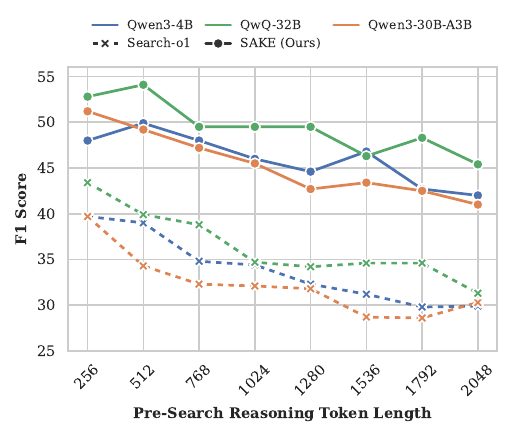}
        \caption{MuSiQue performance across varying pre-search reasoning lengths using oracle search.}
        \label{fig:problem:kid_result_musique}
    \end{subfigure}
    
    \vspace{-5pt}
    \caption{Performance comparison on HotpotQA and MuSiQue datasets across varying pre-search reasoning lengths.}
    \label{fig:problem:kid_results_both}
    \vspace{-10pt}
\end{figure*}
\begin{figure*}[t] 
    \centering
    \begin{subfigure}[b]{0.48\textwidth}
        \centering
        \includegraphics[width=\linewidth]{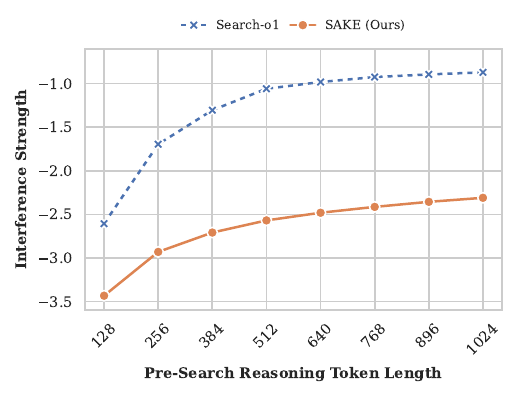}
        \caption{HotpotQA} 
        \label{fig:problem:attn_result_hotpot}
    \end{subfigure}
    \hfill 
    \begin{subfigure}[b]{0.48\textwidth}
        \centering
        \includegraphics[width=\linewidth]{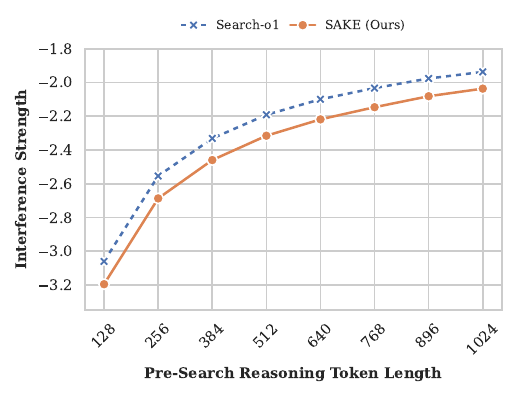}
        \caption{MuSiQue}
        \label{fig:problem:attn_result_musique}
    \end{subfigure}
    
    \vspace{-5pt}
    \caption{Interference strength as a function of pre-search reasoning length computed on HotpotQA (a) and MuSiQue (b) datasets.}
    \label{fig:problem:attn_result_both}
    \vspace{-10pt}
\end{figure*}

Figure~\ref{fig:problem:kid_result_hotpot} and Figure~\ref{fig:problem:kid_result_musique} present the F1 scores on the HotpotQA~\cite{yang2018hotpotqa} and MuSiQue~\cite{trivedi2022musique} datasets, respectively, as a function of pre-search reasoning token length.
Mirroring the trends observed in Figure~\ref{fig:problem:observation}, Search-o1 suffers from distinct performance degradation as the reasoning sequence lengthens in both benchmarks.
In contrast, SAKE successfully mitigates this decline, showcasing superior stability even under extended reasoning contexts.

Furthermore, SAKE consistently maintains a performance advantage over the baseline irrespective of model scale, reinforcing the robustness observed in earlier experiments.
These results support the conclusion that SAKE preserves the semantic integrity of external knowledge, effectively preventing it from being overshadowed by the internal reasoning process.

Extending our analysis of the underlying mechanism across different benchmarks, we further analyze the \textit{interference strength} on HotpotQA and MuSiQue, as illustrated in Figure~\ref{fig:problem:attn_result_hotpot} and Figure~\ref{fig:problem:attn_result_musique}.
Consistent with our findings on the 2WikiMultiHopQA dataset, SAKE maintains significantly lower interference levels compared to Search-o1 as the reasoning length extends.
This suggests that SAKE effectively attenuates the attention-level interference exerted by the reasoning process during knowledge encoding, thereby ensuring that the model allocates sufficient attentional resources to external evidence despite the presence of long reasoning contexts.




\subsection{Performance Analysis across Varying Search Top-$k$}
\begin{figure}[t]
    \centering
    \vspace{-8pt}
    \includegraphics[width=0.46\columnwidth]{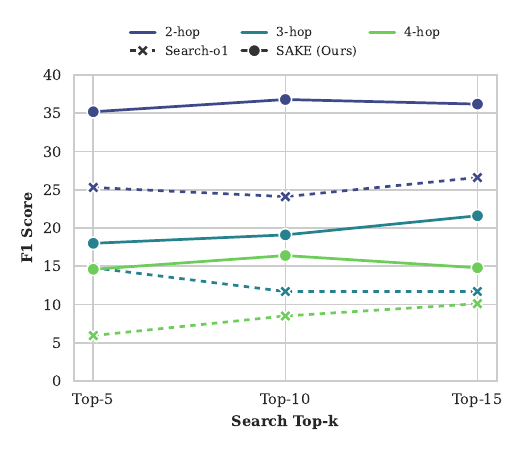}
    \vspace{-6pt}
    \caption{Performance on MuSiQue across different top-k settings and methods.}
    \label{fig:problem:topk_ablation_musique}
    \vspace{-10pt}
\end{figure}
In order to assess the scalability of model performance with respect to the size of retrieved context, we conducted experiments on the MuSiQue dataset by varying the number of retrieved documents (Search Top-$k$).
As illustrated in Figure~\ref{fig:problem:topk_ablation_musique}, SAKE demonstrates robust superiority, consistently outperforming Search-o1 across all $k$ values regardless of the context size.

Intuitively, increasing $k$ raises the likelihood of retrieving the necessary evidence required for the correct answer.
However, Search-o1 fails to exhibit significant performance gains despite this increased recall, suggesting the baseline struggles to fully leverage the expanded external information context.
In contrast, SAKE effectively capitalizes on the richer information available at higher $k$ values.
By ensuring the efficient utilization of retrieved evidence, SAKE translates the increased context into discernible performance improvements.

\subsection{Ablation Study for Knowledge Stack in SAKE}

\begin{table}[h]
    \centering
    \small
    \caption{Results for the ablation study for the knowledge stack of SAKE on MuSiQue.}
    \label{tab:abl_internal}
        \renewcommand{\arraystretch}{1.25}
        \begin{tabular}{lccc}
            \toprule
            \multirow{2}{*}{\textbf{Method}} 
            & \multicolumn{3}{c}{\textbf{MuSiQue}} \\
            \cmidrule(lr{2pt}){2-4}
            
            & 2-hop & 3-hop & 4-hop \\
            \midrule
            Search-o1                                                   &25.3& 14.8& 5.94
\\ 
            SAKE (Ours)                                                 & 35.2& 18.0& \textbf{14.6}
\\
            Flash Stack& \textbf{38.1}& 18.1&14.1
\\ 
            Reversed Stack& 32.8& 19.2&12.9
\\ 
            Brief Stack& 36.4& \textbf{20.7}&10.9\\ 
            \bottomrule
        \end{tabular}
\end{table}

The cumulative knowledge stack proposed in SAKE is characterized by three key properties:  
(1) retrieved knowledge is accumulated across reasoning steps,  
(2) the stack is ordered as $[d_t, \dots, d_1]$, and  
(3) each retrieved document is preserved in its original form without compression or truncation.  
To understand the contribution of each design choice, we conduct an ablation study by modifying these components and evaluating their impact on performance.

Specifically, we consider three variants.  
\textit{(i) Flash Stack} places only the most recent retrieved document at the beginning of the reasoning context, without accumulating past knowledge.  
\textit{(ii) Reversed Stack} arranges retrieved documents in the order $[d_1, \dots, d_t]$, reversing the proposed stacking order.  
\textit{(iii) Brief Stack} includes only a truncated subset of the retrieved content instead of the full documents.

Table~\ref{tab:abl_internal} summarizes the results.  
The Flash Stack exhibits slight performance gains on simpler 2-hop queries but shows a consistent degradation as the task complexity increases to 3-hop and 4-hop settings. This behavior suggests that, while recent evidence may suffice for shallow reasoning, complex multi-hop queries require effective utilization of earlier retrieved knowledge. Since Flash Stack does not preserve prior evidence in the knowledge stack, it fails to support such reasoning.

The Reversed Stack leads to notable performance drops across most settings, except for 3-hop queries. This observation aligns with our analysis in Section~\ref{sec:method}: when the stack is ordered as $[d_1, \dots, d_t]$, the encoding of the current document is more strongly influenced by earlier retrieved content, increasing interference and degrading knowledge integration.

The Brief Stack shows moderate improvements for 2-hop and 3-hop queries but suffers a significant performance drop in the 4-hop setting. This indicates that as reasoning complexity increases, retaining comprehensive external knowledge becomes increasingly important, and aggressive truncation hinders effective multi-step reasoning.

Overall, these results demonstrate that the proposed knowledge stack provides the most stable and robust performance across varying reasoning depths. This ablation study further supports the design choice of SAKE and highlights the importance of structured context management for reliable knowledge integration in agentic reasoning.

\section{Prompts and Code for Experiments}\label{sec:appen:prompts}

Figures~\ref{fig:searcho1_prompt}, \ref{fig:SAKE_prompt}, \ref{fig:RAG_prompt}, and \ref{fig:direct_gen_prompt} present the prompts used for model response generation under each experimental setting.  
Figure~\ref{fig:laj_prompt} shows the prompt employed for LLM-as-a-Judge evaluation on the FRAMES and GAIA benchmarks.  
Figure~\ref{fig:search_function} provides the implementation used to perform web-based search during evaluation on GAIA, where \texttt{GPT-4.1-mini} is utilized as the search backend.

\section{Case Study}\label{sec:appen:prompts}

Figures~\ref{fig:case_study_searcho1_1} and \ref{fig:case_study_searcho1_2} show the problem-solving examples generated by Search-o1 baseline for the queries in MuSiQue dataset. Figures~\ref{fig:case_study_sake_1}, and \ref{fig:case_study_sake_2} show the problem-solving examples generated by our method, SAKE, for the queries in MuSiQue dataset.
As illustrated in the figures, standard search-augmented reasoning systems often fail to properly utilize retrieved evidence, even when the correct answer is explicitly contained in the search results.
This failure arises from Knowledge Integration Decay, where prior reasoning states interfere with the encoding and interpretation of newly retrieved information, leading the model to ignore or distort relevant evidence.
In contrast, when equipped with SAKE, the model effectively preserves the semantic integrity of retrieved knowledge and appropriately incorporates it into subsequent reasoning steps. As a result, the model can correctly recognize answer-relevant evidence from the search results and arrive at the correct final prediction.

\begin{figure}[htbp]
    \small
    \centering
    \begin{mybox}[title=Prompt for Search-o1]
    \hspace*{1em}\{\\
        \hspace*{2em} ``role'': ``system'',\\
        \hspace*{2em} ``content'': ``````

You are a highly capable reasoning assistant with the ability to perform web searches when necessary to answer the user's question accurately.

You have access to a web search tool.

Tool usage format (must be followed exactly):

- To request a search, write: \\
  \verb+<|begin_search_query|>+ your search query \verb+<|end_search_query|>+ \\
- The system will return information in the format:
  \verb+<|begin\_search\_result|>+ ...search results... \verb+<|end\_search\_result|>+ 
  
You can repeat searching up to \textcolor{cyan}{\{MAX\_SEARCH\_LIMIT\}} times. 

General reasoning principles:

1) Decide whether a search is required. \\
\hspace*{1em} - Use search when factual knowledge, verification, or external information is needed. \\
2) Perform focused searches. \\
\hspace*{1em} - Write concise and specific queries targeting the missing information. \\
\hspace*{1em} - Refine queries based on previously returned results if needed. \\
3) Integrate results into your reasoning. \\
\hspace*{1em} - After each search result, extract relevant information. \\
\hspace*{1em} - Decide whether additional searches are necessary. \\
4) Continue reasoning until the question can be answered.

Example (abstract; demonstrates structure only):

Assistant:

\verb+<|begin_search_query|>+ \textless SEARCH\_QUERY\textgreater \verb+<|end_search_query|>+

(System responds)

\verb+<|begin_search_result|>+ ... \textless SEARCH\_RESULTS\textgreater...\verb+<|end_search_result|>+

Assistant continues reasoning, possibly issuing additional searches as needed:

\verb+<|begin_search_query|>+ \textless REFINED\_OR\_NEW\_QUERY\textgreater \verb+<|end_search_query|>+

(System responds)

\verb+<|begin_search_result|>+ ...\textless SEARCH\_RESULTS\textgreater...\verb+<|end_search_result|>+

Assistant continues reasoning...

Important reminders:

- The example above is not a problem to solve and has no fixed number of searches. \\
- Do NOT imitate the example's length or structure mechanically. \\
- Use as many searches as necessary, within the allowed limit. \\
- Always answer the user's actual question, not any placeholder text. \\
- Once you determine that you have sufficient information, produce the answer and stop.

''''''\\
\hspace*{1em}    \}, \\
    \hspace*{1em}\{\\
        \hspace*{2em} ``role'': ``user'',\\
        \hspace*{2em} ``content'': ``````

Please answer the following question. You should provide your final answer in the format \textbackslash\textbackslash boxed\{YOUR\_ANSWER\}.

Question:\\
\textcolor{cyan}{\{Question\}}

''''''\\
\hspace*{1em}    \}
    \end{mybox}
    \caption{Prompt for Search-o1.}
    \label{fig:searcho1_prompt}
\end{figure}
\begin{figure}[htbp]
    \centering
    \begin{mybox}[title=Prompt for SAKE]
    \hspace*{1em}\{\\
        \hspace*{2em} ``role'': ``system'',\\
        \hspace*{2em} ``content'': ``````

You are a highly capable reasoning assistant with the ability to perform web searches when necessary to answer the user's question accurately.

You have access to a web search tool.

Tool usage format (must be followed exactly):

- To request a search, write: \\
  \verb+<|begin_search_query|>+ your search query \verb+<|end_search_query|>+ \\
- The system will return information in the format:
  \verb+<|begin_search_result|>+ ...search results... \verb+<|end_search_result|>+ 
  
You can repeat searching up to \textcolor{cyan}{\{MAX\_SEARCH\_LIMIT\}} times.

\textit{(System prompt is same as Search-o1 prompt) }

''''''\\
\hspace*{1em}    \}, \\
    \hspace*{1em}\{\\
        \hspace*{2em} ``role'': ``user'',\\
        \hspace*{2em} ``content'': ``````

Please answer the following question. You should provide your final answer in the format \textbackslash\textbackslash boxed\{YOUR\_ANSWER\}.

Perform web search based on the following documents. Perform as many searches as necessary.\\
\textcolor{cyan}{\{Search\_Result\}}

Question:\\
\textcolor{cyan}{\{Question\}}

''''''\\
\hspace*{1em}    \}
    \end{mybox}
    \caption{Prompt for SAKE.}
    \label{fig:SAKE_prompt}
\end{figure}
\begin{figure}[htbp]
    \centering
    \begin{mybox}[title=Prompt for standard RAG]
    \hspace*{1em}\{\\
        \hspace*{2em} ``role'': ``user'',\\
        \hspace*{2em} ``content'': ``````

Please answer the following question. You should provide your final answer in the format \textbackslash\textbackslash boxed\{YOUR\_ANSWER\}.

You can refer to the documents related to the question:\\
\textcolor{cyan}{\{Search\_Result\}}

Question:\\
\textcolor{cyan}{\{Question\}}

''''''\\
\hspace*{1em}    \}
    \end{mybox}
    \caption{Prompt for standard RAG.}
    \label{fig:RAG_prompt}
\end{figure}
\begin{figure}[htbp]
    \centering
    \begin{mybox}[title=Prompt for direct generation]
    \hspace*{1em}\{\\
        \hspace*{2em} ``role'': ``user'',\\
        \hspace*{2em} ``content'': ``````

Please answer the following question. You should provide your final answer in the format \textbackslash\textbackslash boxed\{YOUR\_ANSWER\}.

Question:\\
\textcolor{cyan}{\{Question\}}

''''''\\
\hspace*{1em}    \}
    \end{mybox}
    \caption{Prompt for direct generation.}
    \label{fig:direct_gen_prompt}
\end{figure}
\begin{figure}[htbp]
    \centering
    \begin{mybox}[title=Prompt for LLM-as-a-judge]
    \hspace*{1em}\{\\
        \hspace*{2em} ``role'': ``user'',\\
        \hspace*{2em} ``content'': ``````

You are an evaluation assistant. Please determine if the predicted answer is equivalent to the labeled answer. 

Question:\\
\textcolor{cyan}{\{Question\}} 

Labeled Answer:\\
\textcolor{cyan}{\{labeled\_answer\}}

Predicted Answer:\\
\textcolor{cyan}{\{pred\_answer\}}

Are these answers equivalent? Please respond with ''Correct'' if they are equivalent, or "Wrong" if they are not equivalent. Do not include any other text.

''''''\\
\hspace*{1em}    \}
    \end{mybox}
    \caption{Prompt for LLM-as-a-judge.}
    \label{fig:laj_prompt}
\end{figure}
\begin{figure}[htbp]
    \centering
    \begin{mybox}[title=Code for search function]
\noindent
\texttt{def search(query):} \\
\texttt{\hspace*{0.6cm}client = OpenAI(api\_key=OPENAI\_API\_KEY)} \\
\texttt{\hspace*{0.6cm}response = client.responses.create(} \\
\texttt{\hspace*{1.2cm}model="gpt-4.1-mini",} \\
\texttt{\hspace*{1.2cm}tools=[\{"type": "web\_search"\}],} \\
\texttt{\hspace*{1.2cm}tool\_choice="required",} \\
\texttt{\hspace*{1.2cm}input=f"Please perform a web search on the query: \{query\}.\textbackslash n\textbackslash nRetrieve the search} \\
\texttt{\hspace*{1.2cm}results with as much detail as possible.",} \\
\texttt{\hspace*{0.6cm})} \\
\\
\texttt{\hspace*{0.6cm}text = response.output\_text} \\
\texttt{\hspace*{0.6cm}text = re.sub(r"\textbackslash s*\textbackslash (https?://[\textasciicircum\textbackslash )]+ \textbackslash )", "", text)} \\
\texttt{\hspace*{0.6cm}clean\_text = re.sub(r"\textbackslash s*\textbackslash [[\textasciicircum\textbackslash ]]+\textbackslash ]\textbackslash (https?://[\textasciicircum\textbackslash )]+\textbackslash)", "", text)} \\
\texttt{\hspace*{0.6cm}return clean\_text}
    \end{mybox}
    \caption{Code for search function.}
    \label{fig:search_function}
\end{figure}

\begin{figure*}[t]
\centering
\small

\begin{examplepanel}

\begin{sectionbar}
\textbf{Example \#1 from MuSiQue Dataset}
\end{sectionbar}

\vspace{6pt}

\textbf{Question:}\\
\RaggedRight
What other movie did the cast member of Escape from L.A. play a character in?

\vspace{6pt}
\textbf{Labeled Answer:}\\
The Hateful Eight

\vspace{10pt}

\begin{sectionbar}
\textbf{Model Output (Search-o1 Baseline)}
\end{sectionbar}

\vspace{6pt}

Okay, let's tackle this question: "What other movie did the cast member of Escape from L.A. play a character in?" Hmm, first I need to figure out who the cast members of Escape from L.A. are. The user is asking about a specific cast member, but the question doesn't mention which one. Wait, maybe it's referring to a main cast member. Let me think. Escape from L.A. is a 1996 sci-fi action film directed by John Carpenter, starring Kurt Russell as the lead, Snake Plissken. So maybe the user is asking about Kurt Russell's other movies. \\
\medskip
\dots \\
\medskip
But to be accurate, I should search. Let's do the search.
First search: "Kurt Russell Escape from L.A. other movies"
Wait, but the correct way is to search for his filmography. Let's use a search query like "Kurt Russell movies besides Escape from L.A."
Wait, let's use the search tool to get the most accurate answer.
So, I'll generate the search query.\\
\medskip
\textcolor{search_token}{\texttt{<|begin\_search\_query|>}}Kurt Russell Escape from L.A. other movies\textcolor{search_token}{\texttt{<|end\_search\_query|>}}

\medskip
\textcolor{search_token}{\texttt{<|begin\_search\_result|>}} \\
\medskip
\dots Document [5] (Title: The Hateful Eight) The Hateful Eight (often marketed as The H8ful Eight) is a 2015 American western thriller film written and directed by Quentin Tarantino. It stars Samuel L. Jackson, Kurt Russell, \dots \\
\medskip
\textcolor{search_token}{\texttt{<|end\_search\_result|>}}

\medskip
Hmm, looking through the search results, Document [2] mentions that Escape from L.A. is a sequel to Escape from New York, and Kurt Russell stars as Snake Plissken. 
Document [5] mentions Kurt Russell in The Hateful Eight, but that's a different movie. Wait, but the question is asking for "what other movie," so Escape from New York is the direct sequel, so that's the most likely answer. The search results don't explicitly state that Kurt Russell played the same character in Escape from New York, but Document [2] says it's a sequel to Escape from New York. So the other movie would be Escape from New York. \\
\medskip
But let's check if there's another result. \textcolor{error}{Document [5] says Kurt Russell was in The Hateful Eight, but that's a different role.} However, the most notable other movie for him related to Escape from L.A. would be the original Escape from New York.\\
\medskip
\dots \\
\medskip
Let me check if there's any other prominent cast member. For example, Steve Buscemi was in Escape from L.A., but he's known for many movies, like "Fargo," "Reservoir Dogs," etc. But since the question is about "the cast member" (singular), it's more likely referring to the lead, Kurt Russell. So the answer should be Escape from New York. \\
\medskip
Another check: Document [2] says Escape from L.A. is a sequel to Escape from New York, so the other movie is Escape from New York. \\
\medskip
Wait, let's check another search through the search tool. Wait, the search results I got from the first query include Document [2], which states it's a sequel to Escape from New York, so that's the answer. \\
\medskip
\dots \\
\medskip
Therefore, the answer should be Escape from New York. \\
\medskip
\textbf{Final Answer:} \ansbox{Escape from New York}

\end{examplepanel}

\caption{A problem-solving example from the main reasoning process of Search-o1 baseline on the MuSiQue dataset, with special symbols in the search queries and search results highlighted in \textcolor{search_token}{\texttt{blue colors}}.
[\dots] indicates omitted content.}
\label{fig:case_study_searcho1_1}

\end{figure*}
\begin{figure*}[t]
\centering
\small

\begin{examplepanel}

\begin{sectionbar}
\textbf{Example \#1 from MuSiQue Dataset}
\end{sectionbar}

\vspace{6pt}

\textbf{Question:}\\
\RaggedRight
What other movie did the cast member of Escape from L.A. play a character in?

\vspace{6pt}
\textbf{Labeled Answer:}\\
The Hateful Eight

\vspace{10pt}

\begin{sectionbar}
\textbf{Model Output (SAKE)}
\end{sectionbar}

\vspace{6pt}

Okay, let's tackle this question: "What other movie did the cast member of Escape from L.A. play a character in?" Hmm, first I need to figure out who the cast members of Escape from L.A. are. The user is asking about a specific cast member, but the question doesn't mention which one. Wait, maybe it's referring to a main cast member. Let me think. Escape from L.A. is a 1996 sci-fi action film directed by John Carpenter, starring Kurt Russell as the lead, Snake Plissken. So maybe the user is asking about Kurt Russell's other movies.\\
\medskip
\dots \\
\medskip
But to be accurate, I should search. Let's do the search.
First search: "Kurt Russell Escape from L.A. other movies"
Wait, but the correct way is to search for his filmography. Let's use a search query like "Kurt Russell movies besides Escape from L.A."
Wait, let's use the search tool to get the most accurate answer.
So, I'll generate the search query.\\
\medskip
\textcolor{search_token}{\texttt{<|begin\_search\_query|>}}Kurt Russell Escape from L.A. other movies\textcolor{search_token}{\texttt{<|end\_search\_query|>}}

\medskip
\textcolor{search_token}{\texttt{<|begin\_search\_result|>}} \\
\medskip
\dots Document [5] (Title: The Hateful Eight) The Hateful Eight (often marketed as The H8ful Eight) is a 2015 American western thriller film written and directed by Quentin Tarantino. It stars Samuel L. Jackson, Kurt Russell, \dots \\
\medskip
\textcolor{search_token}{\texttt{<|end\_search\_result|>}}

\medskip
Wait, the search results don't include information about Kurt Russell's other movies. The search results provided are the same documents the user listed, but none of them mention Kurt Russell's other films. The only relevant document is [2], which is about Escape from L.A. itself, but it doesn't list his other movies.\\
\medskip
\dots \\
\medskip
Looking at the documents again, in Document [0], there's a mention of Kurt Russell: "Kurt Russell joined the film on the same day O'Brien was in talks. Soon after, John Malkovich was confirmed cast, as a BP representative who fatally underestimates the dangers of working on the rig." This is about Deepwater Horizon (2016), so Kurt Russell was in Deepwater Horizon. So Escape from L.A. (1996) and Deepwater Horizon (2016) are two movies he was in. \\
\medskip
But the question is asking for "other movie," so besides Escape from L.A., Deepwater Horizon is one. But also, in Document [5], The Hateful Eight (2015), Kurt Russell is listed as part of the cast. So he was in The Hateful Eight as well.\\
\medskip
\dots \\
\medskip
Wait, Document [5] says "The Hateful Eight (often marketed as The H8ful Eight) is a 2015 American western thriller film written and directed by Quentin Tarantino. It stars Samuel L. Jackson, Kurt Russell, Jennifer Jason Leigh, Walton Goggins, Demián Bichir, Tim Roth, Michael Madsen, and Bruce Dern..."\\
\medskip
\dots \\
\medskip
Wait, but let's confirm. The question is "What other movie did the cast member of Escape from L.A. play a character in?" The cast member is Kurt Russell, and according to Document [5], he was in The Hateful Eight. So the answer is The Hateful Eight.\\
\medskip
Yes, that's the only other movie mentioned in the provided documents where Kurt Russell is a cast member. So the answer is The Hateful Eight. \\
\medskip
\textbf{Final Answer:} \ansbox{The Hateful Eight}

\end{examplepanel}

\caption{A problem-solving example from the main reasoning process of SAKE on the MuSiQue dataset, with special symbols in the search queries and search results highlighted in \textcolor{search_token}{\texttt{blue colors}}.
[\dots] indicates omitted content.}
\label{fig:case_study_sake_1}

\end{figure*}
\begin{figure*}[t]
\centering
\small

\begin{examplepanel}

\begin{sectionbar}
\textbf{Example \#2 from MuSiQue Dataset}
\end{sectionbar}

\vspace{6pt}

\textbf{Question:}\\
\RaggedRight
In what state is the district where the one who wanted to reform and address the institution behind the religion of Egidio Vagnozzi preached a sermon?

\vspace{6pt}
\textbf{Labeled Answer:}\\
Saxony-Anhalt

\vspace{10pt}

\begin{sectionbar}
\textbf{Model Output (Search-o1 Baseline)}
\end{sectionbar}

\vspace{6pt}

Okay, let's tackle this question. The user is asking about the state where a district is located, which is connected to someone who wanted to reform and address the institution behind the religion of Egidio Vagnozzi. Hmm, first, I need to figure out who Egidio Vagnozzi is.\\
\medskip
\dots \\
\medskip
Wait, the question says "state," which in the US context is a state, but if it's Italy, it's a region. However, the user might be using "state" incorrectly. Let's see.\\
\medskip
\dots \\
\medskip
Wait, let's check another search through the search tool. Let's perform the search to confirm.\\
\medskip
\textcolor{search_token}{\texttt{<|begin\_search\_query|>}}Egidio Vagnozzi reform Catholic Church sermon district state\textcolor{search_token}{\texttt{<|end\_search\_query|>}}

\medskip
\textcolor{search_token}{\texttt{<|begin\_search\_result|>}} \\
\medskip
\dots Document [16] (Title: Wittenberg (district)) Wittenberg is a district () in the east of Saxony-Anhalt, Germany \dots \\
\medskip
\textcolor{search_token}{\texttt{<|end\_search\_result|>}}

\medskip
From the search results, Document [8] states that Egidio Vagnozzi was an Italian Cardinal of the Catholic Church, born in 1906, died in 1980. He was the second president of the Prefecture for the Economic Affairs of the Holy See from 1968 until his death. The Prefecture for the Economic Affairs of the Holy See is based in Vatican City, which is located within Rome, Italy.\\
\medskip
\dots \\
\medskip
\textcolor{error}{But the search results don't mention any specific sermon preached by someone else related to Egidio Vagnozzi.}
However, Document [8] says he was a Cardinal, so he was part of the Catholic Church hierarchy. The person who wanted to reform the Church might be referring to someone else, but the question links it to Egidio Vagnozzi.\\
\medskip
\dots \\
\medskip
However, the user might have intended a US state, but based on the search results, there's no US connection. So the most accurate answer based on the search results is that the district is in Rome, Italy, so the state (region) is Lazio. \\
\medskip
\textcolor{error}{But the question might expect "Rome" or "Italy" as the answer.}
Given the options, and the fact that "state" is used, but in the context of the search results pointing to Rome, I think the intended answer is Rome, Italy, but since the user asked for a "state," and the closest is Lazio, but Lazio is a region, not a state.\\
\medskip
\dots \\
\medskip
Given all this, I think the correct answer is Rome, Italy, but since the user asked for a "state," and assuming they meant the country, but that's not right. Alternatively, the answer is "Italy," but the question says "state." \\
\medskip
\dots \\
\medskip
\textbf{Final Answer:} \ansbox{Rome}

\end{examplepanel}

\caption{A problem-solving example from the main reasoning process of Search-o1 baseline on the MuSiQue dataset, with special symbols in the search queries and search results highlighted in \textcolor{search_token}{\texttt{blue colors}}.
[\dots] indicates omitted content.}
\label{fig:case_study_searcho1_2}

\end{figure*}
\begin{figure*}[t]
\centering
\small

\begin{examplepanel}

\begin{sectionbar}
\textbf{Example \#2 from MuSiQue Dataset}
\end{sectionbar}

\vspace{6pt}

\textbf{Question:}\\
\RaggedRight
In what state is the district where the one who wanted to reform and address the institution behind the religion of Egidio Vagnozzi preached a sermon?

\vspace{6pt}
\textbf{Labeled Answer:}\\
Saxony-Anhalt

\vspace{10pt}

\begin{sectionbar}
\textbf{Model Output (SAKE)}
\end{sectionbar}

\vspace{6pt}

Okay, let's tackle this question. The user is asking about the state where a district is located, which is connected to someone who wanted to reform and address the institution behind the religion of Egidio Vagnozzi. Hmm, first, I need to figure out who Egidio Vagnozzi is.\\
\medskip
\dots \\
\medskip
Wait, the question says "state," which in the US context is a state, but if it's Italy, it's a region. However, the user might be using "state" incorrectly. Let's see.\\
\medskip
\dots \\
\medskip
Wait, let's check another search through the search tool. Let's perform a search for "Egidio Vagnozzi" to see if there's any relevant information.\\
\medskip
\textcolor{search_token}{\texttt{<|begin\_search\_query|>}}Egidio Vagnozzi reform institution religion sermon\textcolor{search_token}{\texttt{<|end\_search\_query|>}}

\medskip
\textcolor{search_token}{\texttt{<|begin\_search\_result|>}} \\
\medskip
\dots Document [16] (Title: Wittenberg (district)) Wittenberg is a district () in the east of Saxony-Anhalt, Germany \dots \\
\medskip
\textcolor{search_token}{\texttt{<|end\_search\_result|>}}

\medskip
Ah! Here's the key. Document [8] mentions "Egidio Vagnozzi (26 February 1906 – 26 December 1980) was an Italian Cardinal of the Catholic Church. He served as the second president of the Prefecture for the Economic Affairs of the Holy See from 1968 until his death, and was elevated to the cardinalate in 1967."\\
\medskip
The "institution behind the religion" would be the Catholic Church, since Egidio Vagnozzi was a Cardinal, part of the Catholic Church. The person who wanted to reform the Catholic Church would be someone like a reformer, but the question says "the one who wanted to reform and address the institution behind the religion of Egidio Vagnozzi preached a sermon."\\
\medskip
\dots \\
\medskip
Wait, looking at Document [14], it mentions Luther's sermon at Wittenberg. Document [14] says: "His final thoughts on Marian devotion and veneration are preserved in a sermon preached at Wittenberg only a month before his death."\\
\medskip
Wittenberg is a district in Saxony-Anhalt, Germany (Document [16] says Wittenberg is a district in east of Saxony-Anhalt, Germany).\\
\medskip
\dots \\
\medskip
Wait, but the question says "the district where... preached a sermon." Document [16] says Wittenberg is a district in Saxony-Anhalt, Germany. So the district is Wittenberg, located in the state of Saxony-Anhalt, Germany. \\
\medskip
But the question asks for the "state," so the answer would be Saxony-Anhalt.\\
\medskip
\dots \\
\medskip
Wait, but let's make sure. The question is phrased as "In what state is the district where the one who wanted to reform and address the institution behind the religion of Egidio Vagnozzi preached a sermon?" The "one" is the reformer (Luther), who preached in Wittenberg (district), which is in Saxony-Anhalt (state).\\
\medskip
\dots \\
\medskip
\textbf{Final Answer:} \ansbox{Saxony-Anhalt}

\end{examplepanel}

\caption{A problem-solving example from the main reasoning process of SAKE on the MuSiQue dataset, with special symbols in the search queries and search results highlighted in \textcolor{search_token}{\texttt{blue colors}}.
[\dots] indicates omitted content.}
\label{fig:case_study_sake_2}

\end{figure*}


\end{document}